\documentclass[runningheads]{llncs}
\usepackage[T1]{fontenc}
\usepackage{graphicx}
\usepackage{multirow}
\usepackage{rotating}
\usepackage[table]{xcolor}

\begin{document}
\title{Applying Transformer-based Text Summarization for Keyphrase Generation\thanks{Supported by the grant of the President of the Russian Federation no. MK-3118.2022.4.}}
%
\titlerunning{Applying Transformer-based Text Summarization for Keyphrase Generation}
%
\author{Anna Glazkova\inst{1}\orcidID{0000-0001-8409-6457} \and
Dmitry Morozov\inst{2}\orcidID{0000-0003-4464-1355}}
%
\authorrunning{A. Glazkova, D. Morozov}
%
\institute{
University of Tyumen, Tyumen, Russia\\
\email{a.v.glazkova@utmn.ru}\\\and
Novosibirsk State University, Novosibirsk, Russia\\
\email{morozowdm@gmail.com}}
\maketitle              
\begin{abstract}
Keyphrases are crucial for searching and systematizing scholarly documents. Most current methods for keyphrase extraction are aimed at the extraction of the most significant words in the text. But in practice, the list of keyphrases often includes words that do not appear in the text explicitly. In this case, the list of keyphrases represents an abstractive summary of the source text. In this paper, we experiment with popular transformer-based models for abstractive text summarization using four benchmark datasets for keyphrase extraction. We compare the results obtained with the results of common unsupervised and supervised methods for keyphrase extraction. Our evaluation shows that summarization models are quite effective in generating keyphrases in the terms of the full-match F1-score and BERTScore. However, they produce a lot of words that are absent in the author’s list of keyphrases, which makes summarization models ineffective in terms of ROUGE-1. We also investigate several ordering strategies to concatenate target keyphrases. The results showed that the choice of strategy affects the performance of keyphrase generation.

\end{abstract}
\section{Introduction}

A list of keyphrases is an important attribute of a scientific text. Keyphrases contain a brief representation of the contents of a text. They help search engines find and systematize papers. A qualitative selection of keyphrases positively affects a paper’s visibility and its number of citations \cite{cite1,cite2}.


Many of the current approaches to keyphrase extraction involve selecting words from the source text, ranking the candidates, and choosing the top $N$. The value of $N$ is determined by the user. The methods of directly extracting keyphrases from the text will produce only those phrases that are explicitly contained in the text. But in practice, keyphrases often represent an abstractive summary of the text. This summary can include hypernyms and paraphrased sentences from the source text. Therefore, studying the applicability of abstractive text summarization methods is a major area of interest within the field of generating multiple keyphrases as a sequence.


Compared with traditional methods for keyphrase extraction, the approaches based on abstractive text summarization have the following properties: a) $N$ is a value determined by the model and is not an input parameter; b) the model takes into account both semantic and syntactic components of the source text; c) not only those words or phrases that are found in the source text can be proposed as keyphrases. To date, few studies have investigated generating keywords using text summarization \cite{cano,swam}. However, the performance of state-of-the-art models based on the transformer architecture has not been closely examined for the task of keyphrase extraction. Moreover, there have been no studies that compare the effectiveness of different ordering strategies for concatenating target keyphrases.


In this paper, we aim to fill this research gap by systematically evaluating transformer-based abstractive text summarization models on several keyphrase extraction benchmarks. We seek to answer the following research questions:
\begin{itemize}
    \item \textbf{RQ1:} Do transformer-based models for abstractive summarization outperform other baselines?
    \item \textbf{RQ2:} What is the effect of different ordering strategies for concatenating target keyphrases?
\end{itemize}

The paper is organized as follows. Section 2 contains a brief review of related works. Next, we describe our methods and experiments in Section 3. In Section 4, we discuss the results. Finally, Section 5 concludes this paper.


\section{Related Work}

The aim of keyphrase extraction is to define a set of phrases that are related to the main topics discussed in a given document \cite{review1}. Up to now, there have been a large volume of published studies presenting unsupervised and supervised approaches to keyphrase extraction.

Unsupervised approaches basically consist of the following stages: selecting candidate words or n-grams in accordance with some characteristics; ranking the candidate words; formatting the keyphrases by selecting the top-ranked words \cite{review}. Unsupervised keyphrase extraction is mainly performed by statistical and graph-based methods. Statistical methods, such as TFIDF, KPMiner \cite{KPMiner}, and YAKE! \cite{Yake}, utilize textual statistical features to define the most important words in the text. The general idea of graph-based methods is to create a document graph consisting of candidate phrases as nodes and their relations as edges. Well-known examples of graph-based methods are TextRank \cite{TextRank}, TopicRank \cite{TopicRank}, and PositionRank \cite{PositionRank}.

Supervised methods for keyphrase extraction include methods based on traditional supervised algorithms as well as deep learning methods. In terms of machine learning, the words in a document are “examples” and the objective of the keyphrase extraction system is to divide the examples into “keyphrases” and “not-keyphrases” \cite{review2}. One of the common keyphrase extraction systems is KEA \cite{Kea} which identifies candidate keyphrases using lexical methods, calculates feature values for each candidate, and uses a Na\"ive Bayes classifier to predict the most probable keyphrases. Binary classification models for keyphrase extraction were also proposed in \cite{sp1,sp2,sp3}. Zhang et al. \cite{rnn} proposed a deep recurrent neural network model to combine keyphrases and context information that jointly process the keyphrase ranking and keyphrase generation task. Meng et al. \cite{KP20K} presented a generative model for keyphrase prediction with an encoder-decoder framework (CopyRNN). Wang et al. \cite{tann} proposed a topic-based adversarial neural network (TANN), which uses the idea of transfer learning. 

To date, the models based on transformer-based models, such as Bidirectional Encoder Representations from Transformers (BERT) \cite{Devlin}, show state-of-the-art results in many natural language processing tasks. Several studies have investigated the use of BERT-based language models for keyphrase extraction. Thus, KeyBERT \cite{KeyBERT} calculates cosine similarity between BERT-embeddings to find the sub-phrases more fully reflecting the content of the document. Some researchers attempted to fine-tune BERT-based models for keyphrase extraction as a classification or sequence labelling task \cite{finetune1,finetune2}. 

A number of studies have examined neural models to generate multiple keyphrases as a sequence \cite{cano,swam}. Chowdhury et al. \cite{BARTfinetuned} showed that fine-tuned BART \cite{BART} can show competitive results in keyphrase generation compared with the existing neural models for keyphrase extraction. The authors ranked the produced keyphrases, selected a fixed number of generated keyphrases per source text, and separately calculated an F1-score for keyphrases present in the source text and Recall for keyphrases absent in the text. Meng et al. \cite{Strategies} explored the effect of different strategies for concatenating target keyphrases with the example of the One2Seq model. They showed that the ordering of concatenating target phrases matters. The best results were achieved where target keyphrases were sorted by their first occurrences in the source text.



\section{Experimental Setup}

\subsection{Datasets}

\begin{table}[]
\addtolength{\tabcolsep}{-0.5pt}
\scriptsize
\centering
\begin{tabular}{|l|l|l|l|l|l|}\hline
\multicolumn{1}{|c}{Characteristic} & \multicolumn{1}{|c}{Krapivin-A} & \multicolumn{1}{|c}{Krapivin-T} & \multicolumn{1}{|c}{KP20K} & \multicolumn{1}{|c}{Inspec} & \multicolumn{1}{|c|}{SemEval2017} \\ \hline
Size & 2294& 2293& 20000&2000&500\\\hline
Domains & CS & CS & CS & CS & \begin{tabular}[c]{@{}l@{}}CS,\\ material science,\\ physics\end{tabular} \\ \hline
Type of texts & abstracts &  texts&  abstracts & abstracts & paragraphs\\ \hline
Avg symbols per text & 1001.74 &  43807.85&  995.85&777.25&1113.66\\ STD &381.37&12565.47&451.32&392.69&310.45\\\hline
Avg tokens per text &  169.06& 8597.63&165.72&127.35&194.99\\ 
STD &68.58&2411.77&76.89&65.03&58.14\\\hline
Avg keyphrases per text &  5.34&  5.34& 5.28&14.11&17.3\\ 
STD &2.77&2.77&3.77&6.41&7\\\hline
Absent keyphrases, \% &  51.3&  18.04& 43.67&43.8&0\\ 
STD &25.99&19.69&28.38&17.83&0\\\hline
\end{tabular}
\caption{\label{font-table} Data statistics. CS - Computer Science. The number of tokens is obtained using NLTK \cite{NLTK}. STD means the standard deviation for the correspondence characteristic.}
\label{table:T1}
\end{table}

We use several English corpora of scientific text for evaluating automatic keyphrase extraction methods. The main characteristics of the datasets are demonstrated in Table \ref{table:T1}. The percentage of absent keyphrases means the proportion of keyphrases from the list of keyphrases that do not appear in the text.

\begin{itemize}
    \item \textbf{Krapivin2009} \cite{Krapivin}, a dataset that contains full papers divided into title, abstract, and text. In this work, we extract keyphrases using abstracts and texts separately (\textbf{Krapivin-A} and \textbf{Krapivin-T} respectively).
    
    \item \textbf{Inspec} \cite{Inspec}, a dataset for keyphrase extraction from scientific abstracts.
    
    \item \textbf{KP20K} \cite{KP20K}, a large-scale scholarly abstracts corpus with 528K abstracts for training, 20K abstracts for validation and 20K abstracts for testing. In this work, we limit ourselves to utilizing only the test set.
    
    \item \textbf{SemEval2017} \cite{Semeval}, a corpus that consists of the paragraphs manually selected from scientific papers among several domains.
\end{itemize}

\subsection{Text Summarization Models}

\begin{figure}[]
    \centering
    \includegraphics[width=1\textwidth]{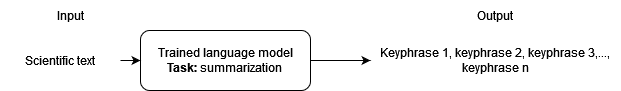}
    \caption{Keyphrase generation with pretrained language models.}
    \label{fig:process}
\end{figure}

For each dataset, we fine-tune two pretrained language models for abstractive text summarization (Figure \ref{fig:process}):

\begin{itemize}
\item \textbf{BART} \cite{BART}, a transformer-based denoising autoencoder for pre-training a seq2seq model. We use BART-base, which has 139 M parameters, 12 layers, and the hidden size of 768;
    \item \textbf{T5} \cite{T5}, an encoder-decoder model pre-trained on a multi-task mixture of unsupervised and supervised tasks and for which each task is converted into a text-to-text format. We utilize T5-small, which has 60M parameters, 12 layers, and the hidden size of 512.
\end{itemize}

We do not fix the number of keyphrases to be generated and thereby this value is determined by the model. This approach can be useful when the number of keyphrases in the training set differs significantly depending on the topic and type of the source text. The other advantage is that the model can independently choose the optimal number of keyphrases based on the training set. However, the model trained in such a manner is not able to generate the exact number of keywords specified during the evaluation stage.

\subsection{Ordering for Concatenating Keyphrases}

We define eight strategies for concatenating target keyphrases, the first six of which were presented in \cite{Strategies}: a) \textbf{No-Sort}, i.e. keyphrases in their original order; b) \textbf{Random}, in random order; c) \textbf{Length}, keyphrases sorted by their length in terms of symbols; d) \textbf{Alpha}, in alphabetical order; e) \textbf{Appear-Pre}, sorted by their first occurrences, absent keyphrases are randomly shuffled and added at the beginning; f) \textbf{Appear-Post}, same as the previous one, but absent keyphrases are added at the end.

In addition to these strategies, we consider the following ways to concatenate keyphrases:

\begin{itemize}
    \item \textbf{Appear-Pre-CC}, same as \textbf{Appear-Pre} but absent and present keyphrases are marked with control codes <absent> and <present> respectively, for example: $\textbf{<absent>}, keyphrase_1, keyphrase_2, ..., keyphrase_{i-1}, \textbf{<present>}, \\keyphrase_i, ..., keyphrase_N$,
    where $N$ is a number of keyphrases. The idea of the use of control codes was introduced in \cite{ctrl}. Control codes represent certain words, sentences, or links which are added to the input of the text generation model to generate coherent outputs. For example, control codes were used for the joined generating titles and short summaries (TLDRs) for scientific papers in \cite{tldr}. The authors appended each source with control codes for short summary and title generation, respectively. This allowed the parameters of the model to learn to generate TLDR or title depending on the control code. In our work, we utilize control codes to distinguish between absent and present keyphrases. 
    
    \item \textbf{Appear-Post-CC}, same as \textbf{Appear-Post} but using control codes.
\end{itemize}

The presented strategies were used during the training phase, i.e. before starting the model training.

\subsection{Baselines}

To evaluate the performance of transformer-based text summarization models for keyphrase generation, we compare them with the following baselines:

\begin{itemize}
    \item \textbf{TFIDF}, a method based on statistical frequencies that expresses the importance of a word in a particular document of a corpus.
    \item \textbf{TopicRank} \cite{TopicRank}, an unsupervised extractive method that extracts the noun phrases and creates top-ranked topic clusters and uses the extracted noun phrases as vertices in a complete graph.
    \item \textbf{YAKE!} \cite{Yake}, an unsupervised multilingual method that utilizes various features, such as term position, term frequency, and others.
    \item \textbf{KEA} \cite{Kea}, a supervised method that extracts candidate keyphrases on the basis of several linguistic features and uses the Na\"ive Bayes algorithm to classify candidate phrases into keyphrases and not.
    \item \textbf{KeyBERT} \cite{KeyBERT}, a method that utilizes document and word embeddings from BERT and cosine similarity to find the sub-phrases in a document that are the most similar to the document itself.
\end{itemize}

\subsection{Evaluation Metrics}

Since previous studies highlighted that different metrics are necessary to capture different aspects of text generation \cite{shen,gurevych}, we adopt three diverse metrics to evaluate the performance of the selected transformer-based text summarization models and baselines: ROUGE-1 \cite{Rouge}, full-match F1-score, and BERTScore  \cite{bertscore}. For those methods that return the predefined number of keyphrases, we computed metrics on the top 5, top 10, and top 15 returned keyphrases.

The ROUGE-1 score calculates the number of matching unigrams between the model-generated text and the reference. To measure ROUGE-1, the keyphrases for each text were combined into a string with a comma as a separator.

The full-match F1-score evaluates the number of exact matches between the original and generated sets of keyphrases. It is calculated as a harmonic mean of precision and recall.

BERTScore utilizes the pre-trained contextual embeddings from BERT-based models and matches words in the source and generated texts using cosine similarity. It has been shown that human judgment correlates with this metric on sentence-level and system-level evaluation. To calculate BERTScore, we use contextual embeddings from RoBERTa-large \cite{roberta}, a modification of BERT that is pretrained using dynamic masking.

\subsection{Implementation Details}

We utilize PKE \cite{pke}, a state-of-the-art open-source Python-based keyphrase extraction toolkit, which makes it possible to implement TFIDF, TopicRank, YAKE!, and KEA. We use the n-gram range of (1,3), which means that produced keyphrases could be unigrams, bigrams, or trigrams. To implement KeyBERT, we use the BERT embeddings produced by the all-MiniLM-L6-v2 model \cite{senttrans} that maps texts to a 384-dimensional dense vector space.

We conduct our experiments on text summarization models using the Transformers library \cite{transformers}. We use 3 epochs to fine-tune models on each dataset, the batch size of 8, the maximum sequence length of 256, and the learning rate of 4e-5. All summarization models are evaluated by three-fold cross-validation. We compute ROUGE-1, F1-score, and BERTScore for each fold separately and then determine the mean value.

\section{Results and Discussion}

To answer \textbf{RQ1}, we compared the results of all considered models in Tables \ref{f1_table}, \ref{r1_table}, and \ref{bs_table} in terms of F1-score, ROUGE-1, and BERTScore, respectively. For baselines, we compute metrics on the top 5, 10, and 15 returned keyphrases. The best result for each dataset is highlighted. The results for Appear-Post, Appear-Pre-CC, and Appear-Post-CC are missed for SemEval2017 because all keyphrases from this dataset are present in the corresponding texts. Therefore, Appear-Pre just produces the list of keyphrases sorted by their first occurrences.

Full-match results on all corpora are reported in Table \ref{f1_table}. Among unsupervised methods, TFIDF shows the highest results on Krapivin-A, Krapivin-T, and KP20K. TopicRank demonstrates the best performance on Inspec and SemEval2017. KEA outperforms other baselines on Krapivin-A (in terms of F1@5), Krapivin-T (F1@10), and Inspec (F1@15). KeyBERT does not show the highest performance on any of the datasets and demonstrates the worst F1-score on Inspec and SemEval2017. Overall, BART achieves the best full-match results on Krapivin-A and Krapivin-T. T5 outperforms other methods on Inspec, KP20K, and SemEval2017.

Table \ref{f1_table} indicates the superiority of abstractive summarization models on the datasets which have a large proportion of absent keyphrases. In particular, the number of summarization models that outperform the best baseline result is higher for the datasets consisting of abstracts, i.e., KP20K (43.67\% of absent keyphrases) -- 9 models, Inspec (43.8\%) -- 8 models, Krapivin-A (51.3\%) -- 3 models, Krapivin-T (18.04\%) -- 1 model, SemEval2017 (0\%) -- 1 model.


\begin{table}[h!]
\addtolength{\tabcolsep}{-0.5pt}
\scriptsize
\begin{tabular}{|cccccccccccccccc|}
\hline
\multicolumn{1}{|c|}{\multirow{2}{*}{Data}} & \multicolumn{3}{c|}{Krapivin-A} & \multicolumn{3}{c|}{Krapivin-T} & \multicolumn{3}{c|}{Inspec} & \multicolumn{3}{c|}{KP20K} & \multicolumn{3}{c|}{SemEval2017} \\ \cline{2-16} 
\multicolumn{1}{|c|}{} & \multicolumn{1}{c|}{\begin{sideways}F1@5\end{sideways}} & \multicolumn{1}{c|}{\begin{sideways}F1@10\end{sideways}} & \multicolumn{1}{c|}{\begin{sideways}F1@15\end{sideways}} & \multicolumn{1}{c|}{\begin{sideways}F1@5\end{sideways}} & \multicolumn{1}{c|}{\begin{sideways}F1@10\end{sideways}} & \multicolumn{1}{c|}{\begin{sideways}F1@15\end{sideways}} & \multicolumn{1}{c|}{\begin{sideways}F1@5\end{sideways}} & \multicolumn{1}{c|}{\begin{sideways}F1@10\end{sideways}} & \multicolumn{1}{c|}{\begin{sideways}F1@15\end{sideways}} & \multicolumn{1}{c|}{\begin{sideways}F1@5\end{sideways}} & \multicolumn{1}{c|}{\begin{sideways}F1@10\end{sideways}} & \multicolumn{1}{c|}{\begin{sideways}F1@15\end{sideways}} & \multicolumn{1}{c|}{\begin{sideways}F1@5\end{sideways}} & \multicolumn{1}{c|}{\begin{sideways}F1@10\end{sideways}} & \multicolumn{1}{c|}{\begin{sideways}F1@15\end{sideways}} \\ \hline
\multicolumn{1}{|c|}{TFIDF} & \multicolumn{1}{c|}{10.8} & \multicolumn{1}{c|}{10.5} & \multicolumn{1}{c|}{9.48} & \multicolumn{1}{c|}{7.85} & \multicolumn{1}{c|}{8.59} & \multicolumn{1}{c|}{8.45} & \multicolumn{1}{c|}{9.76} & \multicolumn{1}{c|}{13.27} & \multicolumn{1}{c|}{14.81} & \multicolumn{1}{c|}{11.12} & \multicolumn{1}{c|}{10.69} & \multicolumn{1}{c|}{9.66} & \multicolumn{1}{c|}{14.86} & \multicolumn{1}{c|}{19.33} & 21.96 \\ \hline
\multicolumn{1}{|c|}{TopicRank} & \multicolumn{1}{c|}{7.61} & \multicolumn{1}{c|}{7.73} & \multicolumn{1}{c|}{7.38} & \multicolumn{1}{c|}{5.4} & \multicolumn{1}{c|}{5.72} & \multicolumn{1}{c|}{5.44} & \multicolumn{1}{c|}{12.01} & \multicolumn{1}{c|}{14.91} & \multicolumn{1}{c|}{16} & \multicolumn{1}{c|}{8.51} & \multicolumn{1}{c|}{8.13} & \multicolumn{1}{c|}{7.54} & \multicolumn{1}{c|}{17.31} & \multicolumn{1}{c|}{22.83} & 24.93 \\ \hline
\multicolumn{1}{|c|}{YAKE!} & \multicolumn{1}{c|}{6.93} & \multicolumn{1}{c|}{8.33} & \multicolumn{1}{c|}{8.1} & \multicolumn{1}{c|}{7.68} & \multicolumn{1}{c|}{8.34} & \multicolumn{1}{c|}{7.79} & \multicolumn{1}{c|}{10.53} & \multicolumn{1}{c|}{13.46} & \multicolumn{1}{c|}{13.96} & \multicolumn{1}{c|}{7.71} & \multicolumn{1}{c|}{8.65} & \multicolumn{1}{c|}{8.58} & \multicolumn{1}{c|}{13.03} & \multicolumn{1}{c|}{18.52} & 21.04 \\ \hline
\multicolumn{1}{|c|}{KEA} & \multicolumn{1}{c|}{10.88} & \multicolumn{1}{c|}{10.47} & \multicolumn{1}{c|}{9.53} & \multicolumn{1}{c|}{8.2} & \multicolumn{1}{c|}{9.19} & \multicolumn{1}{c|}{9.1} & \multicolumn{1}{c|}{9.76} & \multicolumn{1}{c|}{13.17} & \multicolumn{1}{c|}{16.61} & \multicolumn{1}{c|}{7.85} & \multicolumn{1}{c|}{7.49} & \multicolumn{1}{c|}{6.84} & \multicolumn{1}{c|}{14.85} & \multicolumn{1}{c|}{19.42} & 21.72 \\ \hline
\multicolumn{1}{|c|}{KeyBERT} & \multicolumn{1}{c|}{9.46} & \multicolumn{1}{c|}{9.35} & \multicolumn{1}{c|}{8.63} & \multicolumn{1}{c|}{5.54} & \multicolumn{1}{c|}{5.29} & \multicolumn{1}{c|}{4.76} & \multicolumn{1}{c|}{8.62} & \multicolumn{1}{c|}{10.75} & \multicolumn{1}{c|}{11.6} & \multicolumn{1}{c|}{8.27} & \multicolumn{1}{c|}{8.32} & \multicolumn{1}{c|}{7.81} & \multicolumn{1}{c|}{9.47} & \multicolumn{1}{c|}{12.59} & 14.29 \\ \hline
\multicolumn{16}{|c|}{BART} \\ \hline
\multicolumn{1}{|c|}{No-Sort} & \multicolumn{3}{c|}{9.19} & \multicolumn{3}{c|}{5.55} & \multicolumn{3}{c|}{14.45} & \multicolumn{3}{c|}{10.68} & \multicolumn{3}{c|}{10.23} \\ \hline
\multicolumn{1}{|c|}{Random} & \multicolumn{3}{c|}{9.29} & \multicolumn{3}{c|}{5.95} & \multicolumn{3}{c|}{12.14} & \multicolumn{3}{c|}{11.14} & \multicolumn{3}{c|}{12.8} \\ \hline
\multicolumn{1}{|c|}{Length} & \multicolumn{3}{c|}{9.09} & \multicolumn{3}{c|}{5.43} & \multicolumn{3}{c|}{11.77} & \multicolumn{3}{c|}{10.38} & \multicolumn{3}{c|}{13.58} \\ \hline
\multicolumn{1}{|c|}{Alpha} & \multicolumn{3}{c|}{8.71} & \multicolumn{3}{c|}{5.74} & \multicolumn{3}{c|}{9.64} & \multicolumn{3}{c|}{10.72} & \multicolumn{3}{c|}{10.74} \\ \hline
\multicolumn{1}{|c|}{Appear-Pre} & \multicolumn{3}{c|}{10.1} & \multicolumn{3}{c|}{5.17} & \multicolumn{3}{c|}{14.55} & \multicolumn{3}{c|}{11.72} & \multicolumn{3}{c|}{18.22} \\ \hline
\multicolumn{1}{|c|}{Appear-Post} & \multicolumn{3}{c|}{7.59} & \multicolumn{3}{c|}{4.57} & \multicolumn{3}{c|}{13.33} & \multicolumn{3}{c|}{11.36} & \multicolumn{3}{c|}{-} \\ \hline
\multicolumn{1}{|c|}{Appear-Pre-CC} & \multicolumn{3}{c|}{\cellcolor{gray!25}\textbf{11.24}} & \multicolumn{3}{c|}{6.38} & \multicolumn{3}{c|}{15.11} & \multicolumn{3}{c|}{11.76} & \multicolumn{3}{c|}{-} \\ \hline
\multicolumn{1}{|c|}{Appear-Post-CC} & \multicolumn{3}{c|}{9.6} & \multicolumn{3}{c|}{\cellcolor{gray!25}\textbf{9.65}} & \multicolumn{3}{c|}{15.58} & \multicolumn{3}{c|}{11.31} & \multicolumn{3}{c|}{-} \\ \hline
\multicolumn{16}{|c|}{T5} \\ \hline
\multicolumn{1}{|c|}{No-Sort} & \multicolumn{3}{c|}{10.2} & \multicolumn{3}{c|}{6.05} & \multicolumn{3}{c|}{\cellcolor{gray!25}\textbf{22.29}} & \multicolumn{3}{c|}{11.65} & \multicolumn{3}{c|}{16.62} \\ \hline
\multicolumn{1}{|c|}{Random} & \multicolumn{3}{c|}{11} & \multicolumn{3}{c|}{5.9} & \multicolumn{3}{c|}{17.49} & \multicolumn{3}{c|}{\cellcolor{gray!25}\textbf{11.94}} & \multicolumn{3}{c|}{18.62} \\ \hline
\multicolumn{1}{|c|}{Length} & \multicolumn{3}{c|}{10.69} & \multicolumn{3}{c|}{5.31} & \multicolumn{3}{c|}{17.3} & \multicolumn{3}{c|}{10.96} & \multicolumn{3}{c|}{17.51} \\ \hline
\multicolumn{1}{|c|}{Alpha} & \multicolumn{3}{c|}{9.76} & \multicolumn{3}{c|}{6.34} & \multicolumn{3}{c|}{17.43} & \multicolumn{3}{c|}{11.18} & \multicolumn{3}{c|}{18.3} \\ \hline
\multicolumn{1}{|c|}{Appear-Pre} & \multicolumn{3}{c|}{10.93} & \multicolumn{3}{c|}{6.09} & \multicolumn{3}{c|}{19.53} & \multicolumn{3}{c|}{11.43} & \multicolumn{3}{c|}{\cellcolor{gray!25}\textbf{25.39}} \\ \hline
\multicolumn{1}{|c|}{Appear-Post} & \multicolumn{3}{c|}{10.02} & \multicolumn{3}{c|}{6.18} & \multicolumn{3}{c|}{19.04} & \multicolumn{3}{c|}{10.92} & \multicolumn{3}{c|}{-} \\ \hline
\multicolumn{1}{|c|}{Appear-Pre-CC} & \multicolumn{3}{c|}{8.88} & \multicolumn{3}{c|}{4.61} & \multicolumn{3}{c|}{21.24} & \multicolumn{3}{c|}{9.61} & \multicolumn{3}{c|}{-} \\ \hline
\multicolumn{1}{|c|}{Appear-Post-CC} & \multicolumn{3}{c|}{7.38} & \multicolumn{3}{c|}{4.78} & \multicolumn{3}{c|}{18.18} & \multicolumn{3}{c|}{7.53} & \multicolumn{3}{c|}{-} \\ \hline
\end{tabular}
\caption{\label{f1_table} Results (F1-score, \%). The best result for each dataset is highlighted.}
\end{table}

ROUGE-1 scores are reported in Table \ref{r1_table}. TopicRank is the best on Inspec and SemEval2017, KEA demonstrates the highest performance on Krapivin-T, and KeyBERT achieves the best results on Krapivin-A and KP20K. The scores of BART and T5 are quite low on each dataset. To sum up, the results obtained in terms of F1-score and ROUGE-1 show that abstractive text summarization models are relatively successful in predicting full-match keyphrases, but the generated sequence of keyphrases contains a small number of words from the original list of keyphrases. 

\begin{table}[h!]
\addtolength{\tabcolsep}{-0.5pt}
\scriptsize
\begin{tabular}{|cccccccccccccccc|}
\hline
\multicolumn{1}{|c|}{\multirow{2}{*}{Data}} & \multicolumn{3}{c|}{Krapivin-A} & \multicolumn{3}{c|}{Krapivin-T} & \multicolumn{3}{c|}{Inspec} & \multicolumn{3}{c|}{KP20K} & \multicolumn{3}{c|}{SemEval2017} \\ \cline{2-16} 
\multicolumn{1}{|c|}{} & \multicolumn{1}{c|}{\begin{sideways}R1@5\end{sideways}} & \multicolumn{1}{c|}{\begin{sideways}R1@10\end{sideways}} & \multicolumn{1}{c|}{\begin{sideways}R1@15\end{sideways}} & \multicolumn{1}{c|}{\begin{sideways}R1@5\end{sideways}} & \multicolumn{1}{c|}{\begin{sideways}R1@10\end{sideways}} & \multicolumn{1}{c|}{\begin{sideways}R1@15\end{sideways}} & \multicolumn{1}{c|}{\begin{sideways}R1@5\end{sideways}} & \multicolumn{1}{c|}{\begin{sideways}R1@10\end{sideways}} & \multicolumn{1}{c|}{\begin{sideways}R1@15\end{sideways}} & \multicolumn{1}{c|}{\begin{sideways}R1@5\end{sideways}} & \multicolumn{1}{c|}{\begin{sideways}R1@10\end{sideways}} & \multicolumn{1}{c|}{\begin{sideways}R1@15\end{sideways}} & \multicolumn{1}{c|}{\begin{sideways}R1@5\end{sideways}} & \multicolumn{1}{c|}{\begin{sideways}R1@10\end{sideways}} & \multicolumn{1}{c|}{\begin{sideways}R1@15\end{sideways}} \\ \hline
\multicolumn{1}{|c|}{TFIDF} & \multicolumn{1}{c|}{\tiny 27.66} & \multicolumn{1}{c|}{\tiny 29.91} & \multicolumn{1}{c|}{\tiny 29.14} & \multicolumn{1}{c|}{\tiny 21.04} & \multicolumn{1}{c|}{\tiny \tiny 24.84} & \multicolumn{1}{c|}{\tiny 25.21} & \multicolumn{1}{c|}{\tiny 26.55} & \multicolumn{1}{c|}{\tiny 36.45} & \multicolumn{1}{c|}{\tiny 41.45} & \multicolumn{1}{c|}{\tiny 27.9} & \multicolumn{1}{c|}{\tiny 30.11} & \multicolumn{1}{c|}{\tiny 29.3} & \multicolumn{1}{c|}{\tiny 23.67} & \multicolumn{1}{c|}{\tiny 34.92} & \tiny  41.72 \\ \hline
\multicolumn{1}{|c|}{TopicRank} & \multicolumn{1}{c|}{\tiny 24.91} & \multicolumn{1}{c|}{\tiny 25.36} & \multicolumn{1}{c|}{\tiny 24.19} & \multicolumn{1}{c|}{\tiny 19.66} & \multicolumn{1}{c|}{\tiny 21.92} & \multicolumn{1}{c|}{\tiny 21.62} & \multicolumn{1}{c|}{\tiny 31.77} & \multicolumn{1}{c|}{\tiny 40.49} & \multicolumn{1}{c|}{\tiny \cellcolor{gray!25}\textbf{44.09}} & \multicolumn{1}{c|}{\tiny 25.74} & \multicolumn{1}{c|}{\tiny 25.77} & \multicolumn{1}{c|}{\tiny 24.31} & \multicolumn{1}{c|}{\tiny 29.99} & \multicolumn{1}{c|}{\tiny 41.67} & \tiny \cellcolor{gray!25}\textbf{47.78} \\ \hline
\multicolumn{1}{|c|}{YAKE!} & \multicolumn{1}{c|}{\tiny 24.63} & \multicolumn{1}{c|}{\tiny 27.7} & \multicolumn{1}{c|}{\tiny 28.25} & \multicolumn{1}{c|}{\tiny 22.85} & \multicolumn{1}{c|}{\tiny 25.98} & \multicolumn{1}{c|}{\tiny 26.36} & \multicolumn{1}{c|}{\tiny 30.29} & \multicolumn{1}{c|}{\tiny 37.44} & \multicolumn{1}{c|}{\tiny 40.9} & \multicolumn{1}{c|}{\tiny 25.14} & \multicolumn{1}{c|}{\tiny 27.82} & \multicolumn{1}{c|}{\tiny 28.21} & \multicolumn{1}{c|}{\tiny 26.43} & \multicolumn{1}{c|}{\tiny 35.65} & \tiny 41.39 \\ \hline
\multicolumn{1}{|c|}{KEA} & \multicolumn{1}{c|}{\tiny 28.09} & \multicolumn{1}{c|}{\tiny 30.05} & \multicolumn{1}{c|}{\tiny 29.37} & \multicolumn{1}{c|}{\tiny 21.88} & \multicolumn{1}{c|}{\tiny 26.54} & \multicolumn{1}{c|}{\tiny \cellcolor{gray!25} \textbf{27.26}} & \multicolumn{1}{c|}{\tiny 26.49} & \multicolumn{1}{c|}{\tiny 36.03} & \multicolumn{1}{c|}{\tiny 40.88} & \multicolumn{1}{c|}{\tiny 19.86} & \multicolumn{1}{c|}{\tiny 21.36} & \multicolumn{1}{c|}{\tiny 20.89} & \multicolumn{1}{c|}{\tiny 23.51} & \multicolumn{1}{c|}{\tiny 34.4} & \tiny 40.83 \\ \hline
\multicolumn{1}{|c|}{KeyBERT} & \multicolumn{1}{c|}{\tiny 30.22} & \multicolumn{1}{c|}{\cellcolor{gray!25} \textbf{\tiny 31.11}} & \multicolumn{1}{c|}{\tiny 30.49} & \multicolumn{1}{c|}{\tiny 24.78} & \multicolumn{1}{c|}{\tiny 25.16} & \multicolumn{1}{c|}{\tiny 24.01} & \multicolumn{1}{c|}{\tiny 31.01} & \multicolumn{1}{c|}{\tiny 38.33} & \multicolumn{1}{c|}{\tiny 41.93} & \multicolumn{1}{c|}{\tiny 29.68} & \multicolumn{1}{c|}{\tiny \cellcolor{gray!25} \textbf{30.41}} & \multicolumn{1}{c|}{\tiny 29.46} & \multicolumn{1}{c|}{\tiny 26.69} & \multicolumn{1}{c|}{\tiny 36.91} & \tiny  42.77 \\ \hline
\multicolumn{16}{|c|}{BART} \\ \hline
\multicolumn{1}{|c|}{No-Sort} & \multicolumn{3}{c|}{22.69} & \multicolumn{3}{c|}{16.29} & \multicolumn{3}{c|}{36.77} & \multicolumn{3}{c|}{22.9} & \multicolumn{3}{c|}{22.27} \\ \hline
\multicolumn{1}{|c|}{Random} & \multicolumn{3}{c|}{22.88} & \multicolumn{3}{c|}{16.48} & \multicolumn{3}{c|}{31.17} & \multicolumn{3}{c|}{22.97} & \multicolumn{3}{c|}{26.61} \\ \hline
\multicolumn{1}{|c|}{Length} & \multicolumn{3}{c|}{22.28} & \multicolumn{3}{c|}{15.38} & \multicolumn{3}{c|}{30.4} & \multicolumn{3}{c|}{21.6} & \multicolumn{3}{c|}{27.2} \\ \hline
\multicolumn{1}{|c|}{Alpha} & \multicolumn{3}{c|}{21.51} & \multicolumn{3}{c|}{15.72} & \multicolumn{3}{c|}{27.51} & \multicolumn{3}{c|}{22.91} & \multicolumn{3}{c|}{20.23} \\ \hline
\multicolumn{1}{|c|}{Appear-Pre} & \multicolumn{3}{c|}{22.53} & \multicolumn{3}{c|}{16.09} & \multicolumn{3}{c|}{34.38} & \multicolumn{3}{c|}{23.57} & \multicolumn{3}{c|}{38.54} \\ \hline
\multicolumn{1}{|c|}{Appear-Post} & \multicolumn{3}{c|}{21.71} & \multicolumn{3}{c|}{15.27} & \multicolumn{3}{c|}{34.44} & \multicolumn{3}{c|}{17} & \multicolumn{3}{c|}{-} \\ \hline
\multicolumn{1}{|c|}{Appear-Pre-CC} & \multicolumn{3}{c|}{22.22} & \multicolumn{3}{c|}{15.98} & \multicolumn{3}{c|}{34.25} & \multicolumn{3}{c|}{23.52} & \multicolumn{3}{c|}{-} \\ \hline
\multicolumn{1}{|c|}{Appear-Post-CC} & \multicolumn{3}{c|}{21.46} & \multicolumn{3}{c|}{16.47} & \multicolumn{3}{c|}{35.08} & \multicolumn{3}{c|}{17.19} & \multicolumn{3}{c|}{-} \\ \hline
\multicolumn{16}{|c|}{T5} \\ \hline
\multicolumn{1}{|c|}{No-Sort} & \multicolumn{3}{c|}{21.68} & \multicolumn{3}{c|}{14.37} & \multicolumn{3}{c|}{41.51} & \multicolumn{3}{c|}{21.76} & \multicolumn{3}{c|}{27.98} \\ \hline
\multicolumn{1}{|c|}{Random} & \multicolumn{3}{c|}{22.37} & \multicolumn{3}{c|}{14.04} & \multicolumn{3}{c|}{33.52} & \multicolumn{3}{c|}{21.9} & \multicolumn{3}{c|}{33.12} \\ \hline
\multicolumn{1}{|c|}{Length} & \multicolumn{3}{c|}{22.73} & \multicolumn{3}{c|}{13.62} & \multicolumn{3}{c|}{34.27} & \multicolumn{3}{c|}{20.84} & \multicolumn{3}{c|}{30.52} \\ \hline
\multicolumn{1}{|c|}{Alpha} & \multicolumn{3}{c|}{21.66} & \multicolumn{3}{c|}{15.38} & \multicolumn{3}{c|}{32.67} & \multicolumn{3}{c|}{22.02} & \multicolumn{3}{c|}{29.11} \\ \hline
\multicolumn{1}{|c|}{Appear-Pre} & \multicolumn{3}{c|}{22.45} & \multicolumn{3}{c|}{14.94} & \multicolumn{3}{c|}{36.2} & \multicolumn{3}{c|}{21.04} & \multicolumn{3}{c|}{40.34} \\ \hline
\multicolumn{1}{|c|}{Appear-Post} & \multicolumn{3}{c|}{21.07} & \multicolumn{3}{c|}{14.89} & \multicolumn{3}{c|}{36.7} & \multicolumn{3}{c|}{22.13} & \multicolumn{3}{c|}{-} \\ \hline
\multicolumn{1}{|c|}{Appear-Pre-CC} & \multicolumn{3}{c|}{19.1} & \multicolumn{3}{c|}{12.8} & \multicolumn{3}{c|}{39.5} & \multicolumn{3}{c|}{19.48} & \multicolumn{3}{c|}{-} \\ \hline
\multicolumn{1}{|c|}{Appear-Post-CC} & \multicolumn{3}{c|}{18.13} & \multicolumn{3}{c|}{12.47} & \multicolumn{3}{c|}{35.02} & \multicolumn{3}{c|}{18.09} & \multicolumn{3}{c|}{-} \\ \hline
\end{tabular}
\caption{\label{r1_table} Results (ROUGE-1, \%). The best result for each dataset is highlighted.}
\end{table}

The results in terms of BERTScore are shown in Table \ref{bs_table}. Among unsupervised methods, the best scores for all datasets are obtained by TopicRank. KEA and KeyBERT perform worse on all considered text corpora. The results indicate the sustained superiority of BART, which achieves the highest scores for all datasets. The results of T5 differ depending on the data. For instance, T5 shows rather high scores on Inspec but never exceeds the best result obtained by traditional methods on Krapivin-T. In addition, BERTScore also shows that abstractive summarization models mostly outperform traditional methods on the datasets with a large proportion of absent keyphrases, as is the case with F1-score. For example, 15 models of 16 are superior to TopicRank on Inspec and eleven models beat TopicRank on Krapivin-A. On the other hand, only two summarization models of ten outperform the best baseline result on SemEval2017, the proportion of absent keyphrases of which is 0\%. Overall, BERTScore indicates that abstractive summarization models can produce keyphrases that are close in meaning to original keyphrases in terms of token similarity calculated with contextual embeddings.

\begin{table}[h!]
\addtolength{\tabcolsep}{-0.5pt}
\scriptsize
\begin{tabular}{|cccccccccccccccc|}
\hline
\multicolumn{1}{|c|}{\multirow{2}{*}{Data}} & \multicolumn{3}{c|}{Krapivin-A} & \multicolumn{3}{c|}{Krapivin-T} & \multicolumn{3}{c|}{Inspec} & \multicolumn{3}{c|}{KP20K} & \multicolumn{3}{c|}{SemEval2017} \\ \cline{2-16} 
\multicolumn{1}{|c|}{} & \multicolumn{1}{c|}{\begin{sideways}BS@5\end{sideways}} & \multicolumn{1}{c|}{\begin{sideways}BS@10\end{sideways}} & \multicolumn{1}{c|}{\begin{sideways}BS@15\end{sideways}} & \multicolumn{1}{c|}{\begin{sideways}BS@5\end{sideways}} & \multicolumn{1}{c|}{\begin{sideways}BS@10\end{sideways}} & \multicolumn{1}{c|}{\begin{sideways}BS@15\end{sideways}} & \multicolumn{1}{c|}{\begin{sideways}BS@5\end{sideways}} & \multicolumn{1}{c|}{\begin{sideways}BS@10\end{sideways}} & \multicolumn{1}{c|}{\begin{sideways}BS@15\end{sideways}} & \multicolumn{1}{c|}{\begin{sideways}BS@5\end{sideways}} & \multicolumn{1}{c|}{\begin{sideways}BS@10\end{sideways}} & \multicolumn{1}{c|}{\begin{sideways}BS@15\end{sideways}} & \multicolumn{1}{c|}{\begin{sideways}BS@5\end{sideways}} & \multicolumn{1}{c|}{\begin{sideways}BS@10\end{sideways}} & \multicolumn{1}{c|}{\begin{sideways}BS@15\end{sideways}} \\ \hline
\multicolumn{1}{|c|}{TFIDF} & \multicolumn{1}{c|}{\tiny 85.91} & \multicolumn{1}{c|}{\tiny 85.12} & \multicolumn{1}{c|}{\tiny 84.24} & \multicolumn{1}{c|}{\tiny 85.32} & \multicolumn{1}{c|}{\tiny \tiny 85.12} & \multicolumn{1}{c|}{\tiny 84.5} & \multicolumn{1}{c|}{\tiny 83.45} & \multicolumn{1}{c|}{\tiny 84.15} & \multicolumn{1}{c|}{\tiny 84.28} & \multicolumn{1}{c|}{\tiny 85.6} & \multicolumn{1}{c|}{\tiny 84.7} & \multicolumn{1}{c|}{\tiny 83.86} & \multicolumn{1}{c|}{\tiny 83.64} & \multicolumn{1}{c|}{\tiny 84.61} & \tiny  84.73 \\ \hline
\multicolumn{1}{|c|}{TopicRank} & \multicolumn{1}{c|}{\tiny 86.53} & \multicolumn{1}{c|}{\tiny 86.19} & \multicolumn{1}{c|}{\tiny 85.7} & \multicolumn{1}{c|}{\tiny 85.95} & \multicolumn{1}{c|}{\tiny 85.74} & \multicolumn{1}{c|}{\tiny 85.16} & \multicolumn{1}{c|}{\tiny 84.25} & \multicolumn{1}{c|}{\tiny 85.17} & \multicolumn{1}{c|}{\tiny 85.45} & \multicolumn{1}{c|}{\tiny 86.21} & \multicolumn{1}{c|}{\tiny 85.75} & \multicolumn{1}{c|}{\tiny 85.23} & \multicolumn{1}{c|}{\tiny 84.29} & \multicolumn{1}{c|}{\tiny 85.29} & \tiny 85.59 \\ \hline
\multicolumn{1}{|c|}{YAKE!} & \multicolumn{1}{c|}{\tiny 85.28} & \multicolumn{1}{c|}{\tiny 84.74} & \multicolumn{1}{c|}{\tiny 84.14} & \multicolumn{1}{c|}{\tiny 84.98} & \multicolumn{1}{c|}{\tiny 84.73} & \multicolumn{1}{c|}{\tiny 84.19} & \multicolumn{1}{c|}{\tiny 83.64} & \multicolumn{1}{c|}{\tiny 84.1} & \multicolumn{1}{c|}{\tiny 84.21} & \multicolumn{1}{c|}{\tiny 84.88} & \multicolumn{1}{c|}{\tiny 84.24} & \multicolumn{1}{c|}{\tiny 83.72} & \multicolumn{1}{c|}{\tiny 83.5} & \multicolumn{1}{c|}{\tiny 84.2} & \tiny 84.49 \\ \hline
\multicolumn{1}{|c|}{KEA} & \multicolumn{1}{c|}{\tiny 85.9} & \multicolumn{1}{c|}{\tiny 85.06} & \multicolumn{1}{c|}{\tiny 84.2} & \multicolumn{1}{c|}{\tiny 85.49} & \multicolumn{1}{c|}{\tiny 85.41} & \multicolumn{1}{c|}{\tiny 84.83} & \multicolumn{1}{c|}{\tiny 83.44} & \multicolumn{1}{c|}{\tiny 84.1} & \multicolumn{1}{c|}{\tiny 84.21} & \multicolumn{1}{c|}{\tiny 85.56} & \multicolumn{1}{c|}{\tiny 84.66} & \multicolumn{1}{c|}{\tiny 83.82} & \multicolumn{1}{c|}{\tiny 83.59} & \multicolumn{1}{c|}{\tiny 84.51} & \tiny 84.61 \\ \hline
\multicolumn{1}{|c|}{KeyBERT} & \multicolumn{1}{c|}{\tiny 86.43} & \multicolumn{1}{c|}{\tiny 85.46} & \multicolumn{1}{c|}{\tiny 84.72} & \multicolumn{1}{c|}{\tiny 85.45} & \multicolumn{1}{c|}{\tiny 84.21} & \multicolumn{1}{c|}{\tiny 83.33} & \multicolumn{1}{c|}{\tiny 84.39} & \multicolumn{1}{c|}{\tiny 84.64} & \multicolumn{1}{c|}{\tiny 84.56} & \multicolumn{1}{c|}{\tiny 85.72} & \multicolumn{1}{c|}{\tiny 84.71} & \multicolumn{1}{c|}{\tiny 84} & \multicolumn{1}{c|}{\tiny 84.3} & \multicolumn{1}{c|}{\tiny 84.82} & \tiny  84.91 \\ \hline
\multicolumn{16}{|c|}{BART} \\ \hline
\multicolumn{1}{|c|}{No-Sort} & \multicolumn{3}{c|}{87.9} & \multicolumn{3}{c|}{86.59} & \multicolumn{3}{c|}{\cellcolor{gray!25} \textbf{87.72}} & \multicolumn{3}{c|}{86.97} & \multicolumn{3}{c|}{84.42} \\ \hline
\multicolumn{1}{|c|}{Random} & \multicolumn{3}{c|}{88} & \multicolumn{3}{c|}{\cellcolor{gray!25} \textbf{87.01}} & \multicolumn{3}{c|}{86.81} & \multicolumn{3}{c|}{86.98} & \multicolumn{3}{c|}{84.91} \\ \hline
\multicolumn{1}{|c|}{Length} & \multicolumn{3}{c|}{87.85} & \multicolumn{3}{c|}{86.97} & \multicolumn{3}{c|}{86.94} & \multicolumn{3}{c|}{\cellcolor{gray!25}\textbf{87.13}} & \multicolumn{3}{c|}{84.95} \\ \hline
\multicolumn{1}{|c|}{Alpha} & \multicolumn{3}{c|}{87.72} & \multicolumn{3}{c|}{86.68} & \multicolumn{3}{c|}{86.45} & \multicolumn{3}{c|}{87.08} & \multicolumn{3}{c|}{83.91} \\ \hline
\multicolumn{1}{|c|}{Appear-Pre} & \multicolumn{3}{c|}{\cellcolor{gray!25}\textbf{88.09}} & \multicolumn{3}{c|}{86.66} & \multicolumn{3}{c|}{87.36} & \multicolumn{3}{c|}{87.1} & \multicolumn{3}{c|}{\cellcolor{gray!25} \textbf{86.8}} \\ \hline
\multicolumn{1}{|c|}{Appear-Post} & \multicolumn{3}{c|}{87.59} & \multicolumn{3}{c|}{86.64} & \multicolumn{3}{c|}{87.53} & \multicolumn{3}{c|}{85.75} & \multicolumn{3}{c|}{-} \\ \hline
\multicolumn{1}{|c|}{Appear-Pre-CC} & \multicolumn{3}{c|}{86.31} & \multicolumn{3}{c|}{85.43} & \multicolumn{3}{c|}{86.39} & \multicolumn{3}{c|}{86.86} & \multicolumn{3}{c|}{-} \\ \hline
\multicolumn{1}{|c|}{Appear-Post-CC} & \multicolumn{3}{c|}{86.8} & \multicolumn{3}{c|}{86.21} & \multicolumn{3}{c|}{86.73} & \multicolumn{3}{c|}{86.47} & \multicolumn{3}{c|}{-} \\ \hline
\multicolumn{16}{|c|}{T5} \\ \hline
\multicolumn{1}{|c|}{No-Sort} & \multicolumn{3}{c|}{87.3} & \multicolumn{3}{c|}{85.03} & \multicolumn{3}{c|}{86.74} & \multicolumn{3}{c|}{84.84} & \multicolumn{3}{c|}{84.21} \\ \hline
\multicolumn{1}{|c|}{Random} & \multicolumn{3}{c|}{87.02} & \multicolumn{3}{c|}{85.07} & \multicolumn{3}{c|}{85.3} & \multicolumn{3}{c|}{86.79} & \multicolumn{3}{c|}{83.98} \\ \hline
\multicolumn{1}{|c|}{Length} & \multicolumn{3}{c|}{87.29} & \multicolumn{3}{c|}{84.23} & \multicolumn{3}{c|}{85.52} & \multicolumn{3}{c|}{84.11} & \multicolumn{3}{c|}{84.1} \\ \hline
\multicolumn{1}{|c|}{Alpha} & \multicolumn{3}{c|}{86.35} & \multicolumn{3}{c|}{84.51} & \multicolumn{3}{c|}{86.1} & \multicolumn{3}{c|}{85.75} & \multicolumn{3}{c|}{83} \\ \hline
\multicolumn{1}{|c|}{Appear-Pre} & \multicolumn{3}{c|}{86.59} & \multicolumn{3}{c|}{83.57} & \multicolumn{3}{c|}{86.03} & \multicolumn{3}{c|}{86.05} & \multicolumn{3}{c|}{86.43} \\ \hline
\multicolumn{1}{|c|}{Appear-Post} & \multicolumn{3}{c|}{85.32} & \multicolumn{3}{c|}{84.39} & \multicolumn{3}{c|}{85.67} & \multicolumn{3}{c|}{84} & \multicolumn{3}{c|}{-} \\ \hline
\multicolumn{1}{|c|}{Appear-Pre-CC} & \multicolumn{3}{c|}{86.06} & \multicolumn{3}{c|}{83.72} & \multicolumn{3}{c|}{86.31} & \multicolumn{3}{c|}{85.07} & \multicolumn{3}{c|}{-} \\ \hline
\multicolumn{1}{|c|}{Appear-Post-CC} & \multicolumn{3}{c|}{85.05} & \multicolumn{3}{c|}{84.28} & \multicolumn{3}{c|}{85.77} & \multicolumn{3}{c|}{84.25} & \multicolumn{3}{c|}{-} \\ \hline
\end{tabular}
\caption{\label{bs_table} Results (BERTScore, \%). The best result for each dataset is highlighted.}
\end{table}

\begin{figure}[h]
    \centering
    \includegraphics[width=0.8\textwidth]{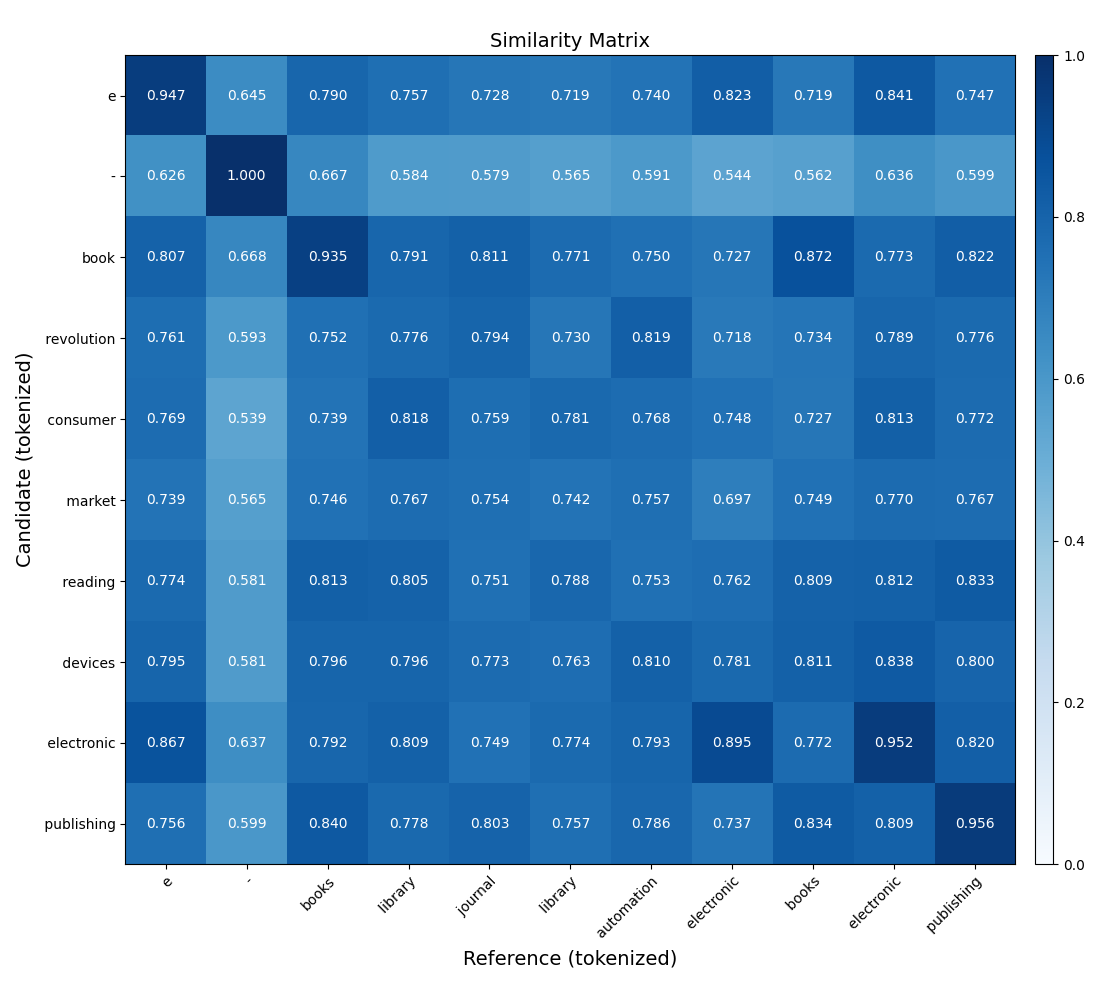}
    \caption{An example of the differences between the metrics. \textbf{Original keyphrases (reference):} e-books, library journal, library automation, electronic books, electronic publishing. \textbf{T5 (candidate):} e-book revolution, consumer market, reading devices, electronic publishing. \textbf{F1-score:} 22.22\%, \textbf{ROUGE-1:} 26.67\%, \textbf{BERTScore:} 88.86\%}
    \label{fig:example}
\end{figure}

Table \ref{examples} shows two examples of generating keyphrases. The first row contains the original list of keyphrases provided by the authors. The rest of the table illustrates keyphrases derived through various methods. For BART and T5, we provide the list of keyphrases produced by the best models. Full matches for the keyphrases from the original list are shown in bold. Exact word matches for the words from the authors' list of keyphrases are underlined. Such examples demonstrate that text summarization models generate more abstractive keyphrases and use fewer repeated words.

Our study demonstrates that different evaluation metrics estimate different aspects of keyphrase generation. For instance, Figure \ref{fig:example} visualizes matching scores between the tokens from original and generated keyphrases obtained by BERTScore. This original example was taken from Inspec and the generated list of keyphrases was produced by T5 (No-Sort). It has an F1-score, which is close to the mean F1-score for this dataset, while the ROUGE-1 score is much lower and BERTScore is higher compared to the mean value. Therefore, the choice of metric is an important step of the research that directly affects the conclusions. In this work, ROUGE-1 indicates the superiority of traditional methods for keyphrase extraction. By contrast, BERTScore defines BART as the most effective model. 

\begin{table}[t!]
\addtolength{\tabcolsep}{-0.5pt}
\scriptsize
\begin{tabular}{|p{1.7cm}|p{4.5cm}|p{5.7cm}|}
\hline
\multicolumn{1}{|c|}{Source} & \multicolumn{1}{|c|}{Example 1} & \multicolumn{1}{|c|}{Example 2} \\ 
\hline
Original & female computer science doctorates, survey of   earned doctorates, information science, computer science education, gender issues & OS porting, application development, consumer operating system, hardware design, operating systems (computers), software portability \\ \hline
TFIDF & \underline{doctorates}, women, completing \underline{doctorates}, \underline{science}, \underline{computer} \underline{science}, 1997, academic year, sed, females, academic & \underline{operating} \underline{system}, \underline{porting}, deliver improved, deliver improved usability, improved usability, high-end portable, high-end portable \underline{consumer}, portable \underline{consumer}, portable \underline{consumer} products, \underline{consumer} products \\ \hline
TopicRank & \underline{doctorates}, \underline{computer}, women, academic year, \underline{science}, percentages, degrees, sed, prior research, females & major complexity, device developer, appropriate \underline{consumer} \underline{operating} \underline{system}, validation, \underline{porting}, problem, support, \underline{system} implementation, trend, real-time \underline{OS} \\ \hline
YAKE! & \underline{education} statistics, national center, \underline{computer} \underline{science}, \underline{computer}, \underline{science}, \underline{doctorates}, degrees, statistics, national, center & portable \underline{consumer} products, high-end portable \underline{consumer}, deliver improved usability, appropriate \underline{consumer} \underline{operating}, \underline{consumer} products, portable \underline{consumer}, appropriate \underline{consumer}, \textbf{consumer operating system}, deliver improved, improved usability \\ \hline
KEA & \underline{doctorates}, women, \underline{science}, \underline{computer} \underline{science}, completing \underline{doctorates}, 1997, academic year, academic, earned, \underline{education} statistics & \underline{operating} \underline{system}, deliver improved, deliver improved usability, improved usability, high-end portable, high-end portable \underline{consumer}, portable \underline{consumer}, portable \underline{consumer} products, \underline{consumer} products, appropriate \underline{consumer} \\ \hline
KeyBERT & \underline{doctorates} \underline{computer}, doctorate \underline{computer}, include \underline{doctorates}, earned \underline{doctorates}, \underline{science} \underline{doctorates}, \underline{computer} \underline{science}, \underline{doctorates}, \underline{doctorates} sed, \underline{doctorates} degrees, \underline{doctorates} include & \textbf{OS porting}, platform \underline{OS}, supported \underline{OS}, \underline{operating} \underline{OS}, \underline{OS}, cation device, portable \underline{consumer}, implementation \underline{porting}, device developer, platform \\ \hline
BART & academic year, \textbf{computer science education}, women \underline{education}, national center for \underline{education} statistics & portable computing, software development, portable \underline{consumer} products, \underline{software}, \textbf{hardware design}, real-time \underline{OS}, \underline{porting}, validation, \underline{software} aspects, portability, performance, user interfaces, user experience, portable products, \textbf{consumer operating system}, \underline{operating} \underline{system} \\ \hline
T5 & \textbf{computer science education}, women, \underline{computer} \underline{science} \underline{doctorates}, academic levels, \textbf{gender issues}, \underline{education} & \underline{porting}, virtual reality, portable \underline{consumer} products, \textbf{consumer operating system}, \underline{OS}, commercially supported \underline{OS}, complete \underline{operating} \underline{system}, real-time \underline{OS}, complex platform \underline{OS}, real-time \underline{operating} \underline{systems}, asynchronous \underline{operating} \underline{systems}, portable devices, mobile computing \\ \hline
\end{tabular}
\caption{\label{examples} Keyphrases extracted by different models. Full matches are bolted, and exact matches are underline.}
\end{table}

To answer \textbf{RQ2}, we estimate the performance growth of the keyphrase generation in comparison with the No-Sort strategy for all other concatenating strategies. The performance growth is calculated as follows:

\begin{equation}
    \frac{R_i-R_{NoSort}}{R_{NoSort}},
\end{equation}

where $R_{NoSort}$ is a performance (F1-score or BERTScore) for the No-Sort strategy, $i$ is a strategy from the set of strategies $I= \{$\textit{Random, Length, Alpha, Appear-Pre, Appear-Post, Appear-Pre-CC, Appear-Post-CC}$\}$. 

\begin{figure}[h!]
\begin{minipage}[h]{1\linewidth}
\center{\includegraphics[width=1\linewidth]{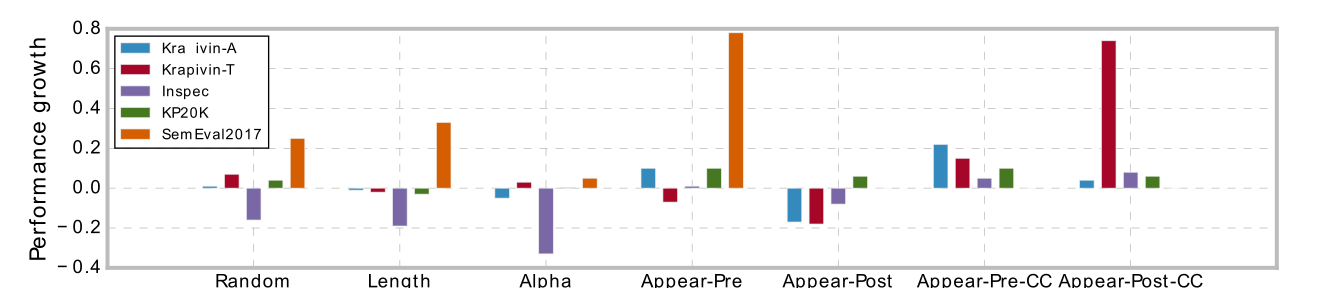} \\ a) BART (F1-score)}
\end{minipage}
\hfill
\begin{minipage}[h!]{1\linewidth}
\center{\includegraphics[width=1\linewidth]{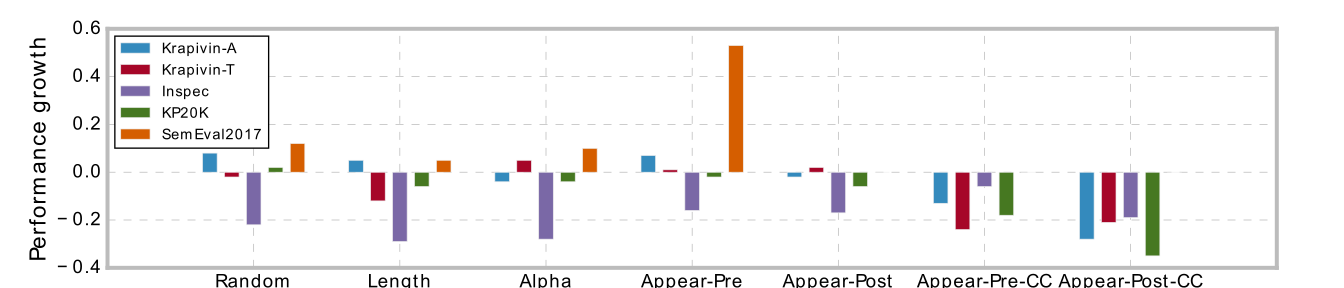} \\ b) T5 (F1-score)}
\end{minipage}
\begin{minipage}[h]{1\linewidth}
\center{\includegraphics[width=1\linewidth]{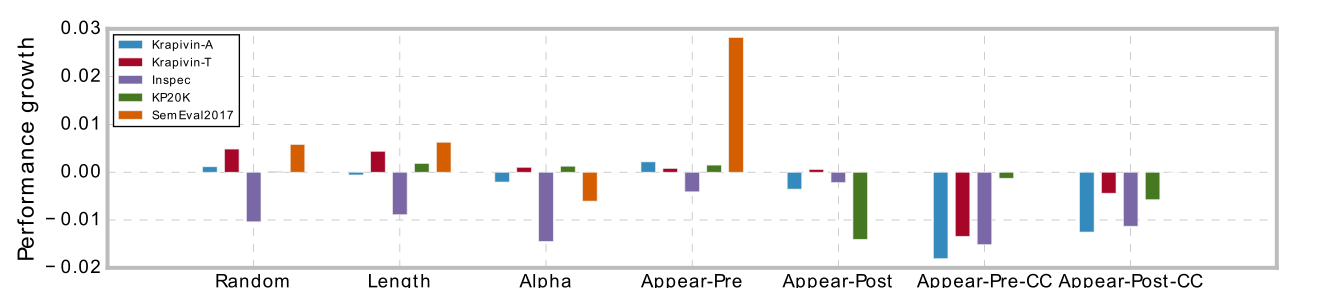} \\ a) BART (BERTScore)}
\end{minipage}
\hfill
\begin{minipage}[h!]{1\linewidth}
\center{\includegraphics[width=1\linewidth]{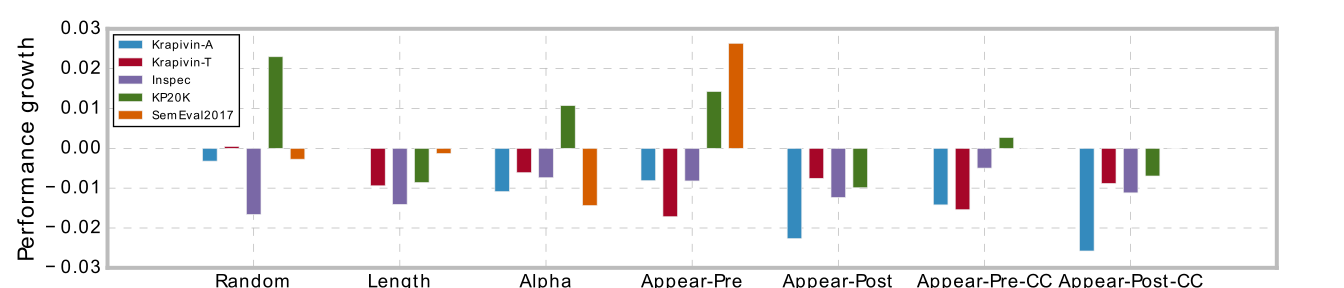} \\ b) T5 (BERTScore)}
\end{minipage}
\caption{Comparing strategies for summarization models.}
\label{fig:image1}
\end{figure}


As shown in Figure \ref{fig:image1}, Random can increase the results, but we have not found stable improvement when using this strategy. Length, Alpha, and Appear-Post demonstrate low performance in most models. As can be seen on all charts, Appear-Pre improves the performance on SemEval2017. Therefore, the ordering is effective for those datasets that does not include absent keyphrases. The results of Appear-Pre-CC and Appear-Post-CC are opposite for BART (F1-Score) and other cases. The use of control codes improves the full-match results of BART on all corpora. In all other considered cases, control codes appeared to be ineffective.

\begin{figure}[h!]
    \centering
    \includegraphics[width=0.75\textwidth]{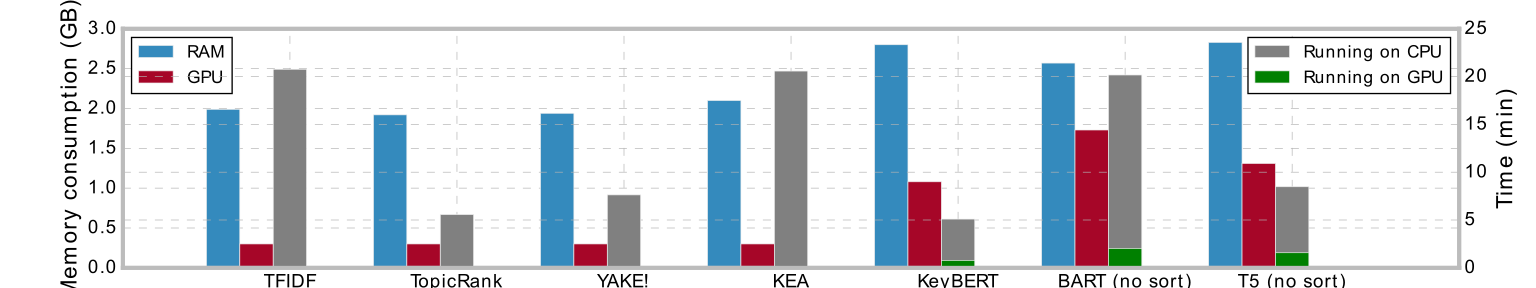}
    \caption{Resource usage on the example of SemEval2017.}
    \label{fig:memory}
\end{figure}

To illustrate the resource usage for keyphrase generation by different models, we show the time and memory consumption in Figure \ref{fig:memory} on the example of SemEval2017. The time and memory usage for training was not included in the calculation. In other words, we loaded the trained model, generated keyphrases for all texts in the dataset, and measured the indicators using the Python and Google Colaboratory tools. The figure shows that memory usage is expectedly higher for transformer-based models than for traditional methods. The time consumption during running on CPU is the highest for TFIDF, KEA, and BART. However, the use of GPU for transformers allows a reduction of the generation time to less than one minute for KeyBERT and approximately two minutes for BART and T5.

\section{Conclusion}

In this paper, we explored the effectiveness of abstractive summarization models based on transformer architecture for the task of predicting keyphrases for scientific texts. We performed an extensive evaluation of unsupervised and supervised models for keyphrase extraction and compared several ordering strategies for concatenating keyphrases on several datasets. Our results showed some pros and cons of the use of transformer-based summarization models for keyphrase extraction. First, we obtained promising results in terms of the full-match F1-score and BERTScore, but ROUGE-1 indicates the superiority of traditional methods for keyphrase extraction. Second, we indicated that summarization models are more competitive in generating keyphrases that are not explicitly presented in the source text. Finally, we demonstrated that some ordering strategies provide better results in keyphrase generation, while others decrease the performance. 
%
%
%
%

\end{document}